\documentclass[conference]{IEEEtran}
\IEEEoverridecommandlockouts
\usepackage{cite}
\usepackage{amsmath,amssymb,amsfonts}
\usepackage{graphicx}
\usepackage{textcomp}
\usepackage{xcolor}
\usepackage{comment}
\usepackage{hyperref}
\usepackage{graphicx}
\usepackage{pgfplots}
\usepackage{subcaption}
\usepackage{ifthen}
\usepackage{algorithm}
\usepackage{algpseudocode}
\usepackage{multirow}
\usepackage{makecell}
\usepackage{enumitem}
\usepackage[acronym,toc,nonumberlist]{glossaries} 
\usepackage{booktabs}
\usetikzlibrary{matrix,calc}

\newacronym{ffp}{FFP}{Fuzzy Fingerprint}
\newacronym{nlp}{NLP}{Natural Language Processing}
\newacronym{nn}{NN}{Neural Network}
\newacronym{cnn}{CNN}{Convolutional Neural Network}
\newacronym{tsi}{TSI}{Textual Speaker Identification}

\DeclareMathOperator*{\softmax}{softmax}

\def\BibTeX{{\rm B\kern-.05em{\sc i\kern-.025em b}\kern-.08em
    T\kern-.1667em\lower.7ex\hbox{E}\kern-.125emX}}
\begin{document}

\title{Speaker Fuzzy Fingerprints: Benchmarking Text-Based Identification in Multiparty Dialogues
\thanks{This work was supported by national funds through FCT, Fundação para a Ciência e a Tecnologia, under project UIDB/50021/2020 (DOI:10.54499/UIDB/50021/2020) and grant 2022.10640.BD, and by the Recovery and Resilience Plan (RRP) and Next Generation EU European Funds through project C644865762-00000008 Accelerat.AI.}
}

%\author{\IEEEauthorblockN{Anonymous Authors}}
\author{\IEEEauthorblockN{Rui Ribeiro}
\IEEEauthorblockA{\textit{INESC-ID} \\
\textit{Instituto Superior Técnico}\\
\textit{Universidade de Lisboa}\\
Lisbon, Portugal \\
{\tt \scriptsize rui.m.ribeiro@inesc-id.pt}}
\and
\IEEEauthorblockN{Luísa Coheur}
\IEEEauthorblockA{\textit{INESC-ID} \\
\textit{Instituto Superior Técnico}\\
\textit{Universidade de Lisboa}\\
Lisbon, Portugal \\
{\tt \scriptsize luisa.coheur@inesc-id.pt}}
\and
\IEEEauthorblockN{Joao P. Carvalho}
\IEEEauthorblockA{\textit{INESC-ID} \\
\textit{Instituto Superior Técnico}\\
\textit{Universidade de Lisboa}\\
Lisbon, Portugal \\
{\tt \scriptsize joao.carvalho@inesc-id.pt}}
}

\maketitle

\begin{abstract}
Speaker identification using voice recordings leverages unique acoustic features, but this approach fails when only textual data is available.
Few approaches have attempted to tackle the problem of identifying speakers solely from text, and the existing ones have primarily relied on traditional methods.
In this work, we explore the use of fuzzy fingerprints from large pre-trained models to improve text-based speaker identification. We integrate speaker-specific tokens and context-aware modeling, demonstrating that conversational context significantly boosts accuracy, reaching 70.6\% on the \emph{Friends} dataset and 67.7\% on the \emph{Big Bang Theory} dataset. Additionally, we show that fuzzy fingerprints can approximate full fine-tuning performance with fewer hidden units, offering improved interpretability. Finally, we analyze ambiguous utterances and propose a mechanism to detect speaker-agnostic lines. Our findings highlight key challenges and provide insights for future improvements in text-based speaker identification.
\end{abstract}


\section{Introduction}

Speaker identification from voice recordings has been a widely active topic in the Speech Recognition area \cite{VOXCELEB2,WAV2VEC2,BIAS-SPEAKER,TEIXEIRA-SPEECH}, where the goal is to identify the speaker by using the information contained in the speaker's speech signal.
When speaking, an interlocutor comprises features such as the properties of the speaking style that are proper of their idiosyncratic attributes and can be exploited as acoustic features for speaker recognition models. 
However, when speech is transcribed into text, these acoustic features are lost, removing valuable speaker-specific cues. This limitation makes it significantly harder to distinguish speakers based solely on linguistic patterns. Furthermore, in many real-world scenarios, only textual data is available, such as in movie scripts, chat logs, and historical records, necessitating effective speaker identification methods that rely purely on text.

Only a few approaches have attempted to identify the speaker given textual conversational data. For instance, Ma et al. \cite{TEXT-SPEAKER-CNN} trained a \gls*{cnn} to identify textual speakers from multiparty dialogues extracted from the \textit{Friends} TV Show, demonstrating that distinct speaker styles can be captured from text alone and achieving significant improvements over traditional methods. However, with recent advancements in Natural Language Processing, these results are no longer state-of-the-art. Beyond CNN-based approaches, other methodologies have also been explored. Kundu et al. \cite{SPEAKER-FILM-DIALOGUES} applied traditional machine learning techniques to detect speaker boundaries and changes in automatically transcribed dialogues, while Su and Zhou \cite{SPEAKER-CLUSTERING} proposed a speaker clustering model that groups utterances without explicit speaker annotations. While these methods highlight the feasibility of textual speaker identification, the task remains inherently difficult due to the frequent use of generic and short utterances and the stylistic overlap between speakers.

In this work, we explore new mechanisms to enhance speaker identification by better capturing speaker-specific linguistic patterns and contextual dependencies. To this end, we integrate speaker-specific tokens and context-aware modeling to improve identification performance and reduce ambiguity.
Additionally, to ensure better generalization across different domains, we explore a new corpus derived from the \textit{Big Bang Theory} TV series, complementing existing datasets. 
Beyond improving classification accuracy, we also investigate the use of the Fuzzy Fingerprint framework to detect ambiguous or generic utterances that lack strong speaker-specific cues.

Our contributions to textual speaker identification include:
\begin{itemize}
    \item We explore a new corpus of textual dialogues from the \textit{Big Bang Theory} that complements the \textit{Friends} dataset, expanding the range of conversational styles.
    \item We benchmark both datasets\footnote{\url{https://github.com/ruinunca/speaker-fuzzy-fingerprints}} using a recent technique that leverages fuzzy fingerprints from language models (specifically RoBERTa variants), analyzing the impact of conversational context length.
    \item We explore the use of the Fuzzy Fingerprint framework to reduce the number of hidden units while maintaining performance comparable to fully fine-tuned models, as well as to investigate the impact of speaker-agnostic utterances.
\end{itemize}

%Our results confirm that textual cues alone provide valuable signals for speaker identification, particularly when enriched with context. However, challenges remain, as some utterances are inherently ambiguous and difficult to attribute to a single speaker. By sharing the datasets' splits and code publicly available\footnote{\url{https://link-to-repository}}, we aim to facilitate further advancements in textual speaker identification, offering a reliable alternative for scenarios where speech data is unavailable.

%------------------------------------
\section{Related Work}

Although many approaches to speaker recognition have traditionally relied on acoustic characteristics extracted from speech signals~\cite{VOXCELEB2,WAV2VEC2,BIAS-SPEAKER,TEIXEIRA-SPEECH}, text-centered methods are receiving growing attention for scenarios where audio data are unavailable. One of the earlier efforts to leverage purely textual cues focused on identifying speaker shifts in automatically transcribed dialogs. 
For instance, Kundu \textit{et al.}~\cite{SPEAKER-FILM-DIALOGUES} employed standard machine learning algorithms (such as K-Nearest Neighbors and Naive Bayes) to detect speaker transitions. % by analyzing basic textual attributes at sentence boundaries, laying the groundwork for more advanced text-based speaker identification.
Ma \textit{et al.}~\cite{TEXT-SPEAKER-CNN} devised a CNN tailored for speaker classification in multiparty interactions drawn from the \textit{Friends} TV series. % Their results showed that distinctive linguistic habits can be captured strictly from text and that even nuanced language patterns help surpass traditional rule-based or statistical methods.
More recently, Su and Zhou~\cite{SPEAKER-CLUSTERING} proposed an unsupervised paradigm that clusters utterances without relying on speaker labels. By integrating a pre-trained language model at the utterance level with a pairwise similarity matrix, the authors demonstrated that natural groupings of utterances can accurately align with speaker identities, further establishing the effectiveness of textual features in speaker recognition tasks.

%Speaker identification has traditionally relied on acoustic features derived from speech data \cite{VOXCELEB2,WAV2VEC2,BIAS-SPEAKER,TEIXEIRA-SPEECH}. However, although text-based approaches are crucial when audio data is unavailable, only a few works have attempted to tackle this challenge.
%\cite{SPEAKER-FILM-DIALOGUES} explored traditional methods such as the K-Nearest Neighbor Algorithm and Naive Bayes Classifier focused on sentence boundary detection and speaker change detection in automatically transcribed texts.
%\cite{TEXT-SPEAKER-CNN} proposed a CNN model for text-based speaker identification on multiparty dialogues from the TV show \textit{Friends}. 
%This work discusses the potential of text-based approaches by leveraging distinct styles of speakers, achieving significant improvements over baseline models. 
%Similarly, \cite{SPEAKER-CLUSTERING} introduced a speaker clustering model for textual dialogues, which groups utterances without speaker annotations. 
%By leveraging a pre-trained language model and an utterance-level pairwise matrix, the authors effectively captured semantic and relational aspects of dialogues, demonstrating the efficacy of clustering in speaker identification tasks.

Beyond conventional machine learning and deep learning approaches, alternative representation techniques may provide complementary advantages for text-based speaker identification. One such method is the Fuzzy Fingerprint framework, which has been utilized as an efficient technique for generating compact and distinctive representations of large datasets~\cite{homem2011web}.
This method has demonstrated its utility in various applications, including authorship attribution, topic classification, and emotion detection~\cite{pereira2023fuzzy, RUI-FFP-GENETIC-IFSA, RUI-FFP-AUTHORSHIP}.
Fuzzy fingerprints are formed by accumulating feature activations (e.g., word occurrences or neural embeddings) from all training samples of a particular class, ordering these features from highest to lowest occurrence, and retaining the most salient subset as the class’s core representation. 
In the end, a fuzzy fingerprint is a fuzzy set in the discrete universe of the used features~\cite{RUIFUZZY}.
%A library of these class-level fuzzy fingerprints is then used to quickly match any new instance to the class with the strongest similarity score, effectively leveraging the membership-based intersection of features to minimize collisions between classes. 

%------------------------------------
\section{Corpora}

%We utilize two datasets from popular TV sitcoms: the Friends dataset and the Big Bang Theory dataset. These datasets capture various interactions among main characters in a multiparty setting, making them well-suited for the speaker identification task.
%These datasets provide a consistent source of conversational data which is essential for character identification and dialogue analysis tasks.

\subsection{Friends Corpus}

We use the \textit{Friends} dataset introduced by \cite{chen2016character}. Each season contains multiple episodes, and each episode is comprised of separate scenes. The scenes in an episode are divided into turns, containing the annotation of the speakers.

\begin{figure}[ht!]
    \centering
    \begin{tikzpicture}[scale=0.8]
        \begin{axis}[
            ybar,
            symbolic x coords={O, RO, RA, J, C, P, M},
            xtick=data,
            ylabel={Relative Frequency (\%)},
            xlabel={Character},
            bar width=15pt,
            ymin=0,
            ymax=20,
            nodes near coords,
        ]
            \addplot[fill=red!50] coordinates {
                (O, 16.29)
                (RO, 15.00)
                (RA, 15.27)
                (J, 13.42)
                (C, 13.89)
                (P, 12.30)
                (M, 13.84)
            };
        \end{axis}
    \end{tikzpicture}
    \caption{Distribution of Turns per Speaker for the \textit{Friends} Corpus (O - Other; RO - Ross; RA - Rachel; J - Joey; C - Chandler; P - Phoebe; M - Monica).}
    \label{figure:friends_speaker_distribution}
\end{figure}
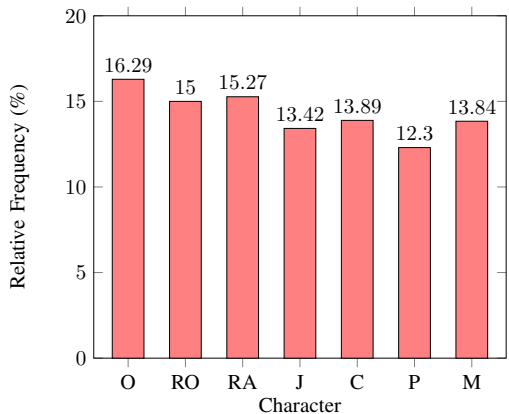

\begin{table}[ht!]
    \centering
    \begin{tabular}{l|c|c|c}
        \hline
        \textbf{Statistic} & \textbf{Train} & \textbf{Valid} & \textbf{Test} \\
        \hline
        Total Scenes & 2268 & 332 & 288 \\
        Total Turns & 43799 & 6343 & 6231 \\
        Mean Sentence Length & 10.18 & 9.92 & 10.17 \\
        Std Deviation of Sentence Length & 10.59 & 10.16 & 10.32 \\
        Mean Scene Length & 19.31 & 19.11 & 21.64 \\
        Std Deviation of Scene Length & 15.74 & 14.04 & 18.06 \\
        \hline
    \end{tabular}
    \caption{Statistics of train, validation, and test splits of the \textit{Friends} dataset.}
    \label{tab:friends-splits}
\end{table}

In total, this corpus consists of 3,107 scenes and 61,676 turns. The distribution of utterances per speaker is illustrated in Figure \ref{figure:friends_speaker_distribution}. 
We only consider the 6 main characters for the labels, while the other characters are considered as the \textit{Other} label. 
The percentages for major speakers are fairly consistent; however, the \textit{Other} speaker category has a larger percentage in the dataset than any individual major speaker.
Following \cite{TEXT-SPEAKER-CNN}, we consider season 7 as the validation set, season 8 as the test, and the remaining as training data.
Table \ref{tab:friends-splits} shows the statistics for the train, validation, and test splits.

\subsection{Big Bang Theory Corpus}
\begin{table}[ht!]
    \centering
    \begin{tabular}{l|c|c|c}
        \hline
        \textbf{Statistic} & \textbf{Train} & \textbf{Valid} & \textbf{Test} \\
        \hline
        Total Scenes & 2280 & 285 & 285 \\
        Total Turns & 41247 & 5237 & 5072 \\
        Mean Sentence Length & 11.36 & 11.51 & 11.05 \\
        Std Deviation of Sentence Length & 10.68 & 10.88 & 9.99 \\
        Mean Scene Length & 18.09 & 18.38 & 17.80 \\
        Std Deviation of Scene Length & 13.07 & 13.18 & 10.88 \\
        \hline
    \end{tabular}
    \caption{Statistics of train, validation, and test splits of the \textit{Big Bang Theory} dataset.}
    \label{tab:big-bang-splits}
\end{table}

We leverage the \textit{Big Bang Theory} dataset\footnote{\url{https://www.kaggle.com/datasets/mitramir5/the-big-bang-theory-series-transcript}} from online transcripts of the sitcom \textit{Big Bang Theory}\footnote{\url{https://bigbangtrans.wordpress.com/}}. 
The resulting dataset provides a comprehensive collection of dialogues from the show's episodes.
Similar to the \textit{Friends} dataset, the transcripts are structured by episodes, scenes, and utterances.

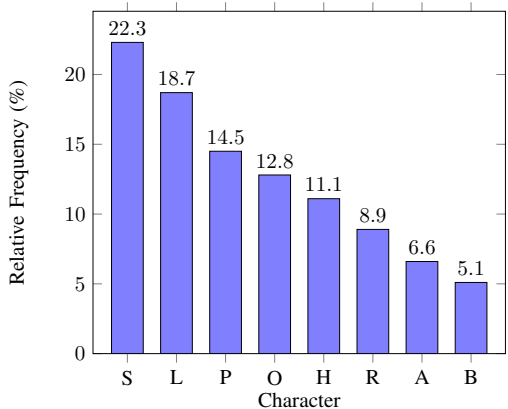
\begin{figure}[ht]
    \centering
    \begin{tikzpicture}[scale=0.8]
        \begin{axis}[
            ybar,
            symbolic x coords={S, L, P, O, H, R, A, B},
            xtick=data,
            ylabel={Relative Frequency (\%)},
            xlabel={Character},
            bar width=15pt,
            ymin=0,
            nodes near coords,
        ]
            \addplot[fill=blue!50] coordinates {
                (S, 22.3)
                (L, 18.7)
                (P, 14.5)
                (O, 12.8)
                (H, 11.1)
                (R, 8.9)
                (A, 6.6)
                (B, 5.1)
            };
        \end{axis}
    \end{tikzpicture}
    \caption{Turn distribution among characters in the \textit{Big Bang Theory} dataset (S - Sheldon; L - Leonard; P - Penny; O - Other; H - Howard; R - Raj; A - Amy; B - Bernadette).}
    \label{fig:turn-distribution}
\end{figure}

The turn distribution among characters in the \textit{Big Bang Theory} dataset is illustrated in Figure \ref{fig:turn-distribution}. 
As in the \textit{Friends} corpus, we also make the distinction between main characters and \textit{Other} characters.
%Sheldon has the highest percentage of turns at 22.27\%, followed by Leonard with 18.69\%, and Penny with 14.50\%. Other characters, including Howard, Raj, Bernadette, and Amy, have lower percentages, reflecting the distribution of dialogue among the main cast.
For this corpus, we randomly divide the data into training, validation, and test splits instead of choosing specific seasons for the splits as in the \textit{Friends} dataset.
The reason is that using a particular season could result in a bias toward a specific vocabulary, as characters may suffer changes in their behaviors between seasons.
%This results in two different manners to evaluate the textual speaker identification task.
Table \ref{tab:big-bang-splits} shows the statistics of the train, validation, and test splits of the \textit{Big Bang Theory} dataset.
%Both datasets are essential for training and evaluating our models in character identification and dialogue understanding tasks, providing a diverse and extensive collection of conversational data from two iconic TV sitcoms.

%------------------------------------
\section{Experiments}

\subsection{Fuzzy Fingerprint from Large Language Models}
\label{section:proposed-method}

We follow the approach from~\cite{RUIFUZZY} and merge the Fuzzy Fingerprint framework with large pre-trained encoders. 
Specifically, given a model \(\mathcal{M}^{E}\) (e.g., RoBERTa) that yields a hidden representation \(h \in \mathbb{R}^{M}\) for an input text \(T\), we first fine-tune \(\mathcal{M}^{E}\) on the target dataset to learn a classification function \(p(c \mid T) = \softmax(Wh)\), where \(W \in \mathbb{R}^{|C|\times M}\). Once the model is fine-tuned, we follow four main steps to build fuzzy fingerprints:

\paragraph{Summation of Activations}
For each class \(c\), we gather all texts labeled with \(c\), pass them through \(\mathcal{M}^{E}\), and sum the absolute values of the resulting hidden states. This generates a single accumulated vector \(V_{c} \in \mathbb{R}^M\), where each entry \(i\) denotes the cumulative activation of hidden unit \(i\).

\paragraph{Ranking Hidden Units}
We rank these hidden units for each class from most to least activated, effectively identifying which parts of the model's representation are most indicative of class \(c\). By focusing on the top-$k$ units, we create a compact signature for each class.

\paragraph{Membership Function}
Following \cite{RUIFUZZY}, we apply a Pareto-based membership function to highlight the top-$k$ ranked units. Each hidden unit $i$ in class $c$'s vector is assigned a membership value $\mu_{i}(\Phi_c)$, which is nonzero only for the most activated $k$ units, yielding a fuzzy fingerprint \(\Phi_c\).

\paragraph{Fuzzy Fingerprint Library}
We compile these class-specific fuzzy fingerprints into a \emph{Fuzzy Fingerprint Library}. Given a new text \(T\), we derive its fuzzy fingerprint \(\phi_T\) by summing the absolute values of its hidden states and ranking them accordingly. 
Classification occurs by measuring the fuzzy similarity of the sample's fuzzy fingerprint \(\phi_T\) to each class fuzzy fingerprint \(\Phi_c\):
\begin{equation}
\label{eq:similarity}
\mbox{similarity}(\phi_T, \Phi_c) 
= \sum_{v=1}^{M} \frac{\top(\mu_v(\phi_T), \mu_v(\Phi_c))}{N},
\end{equation}
where $N$ is a normalization constant (often $k$) and $\top$ is a T-norm (e.g., the $\min$ norm). 
The similarity function is the aggregation of the intersection of the fuzzy set representing the class with the fuzzy set representing the text \(T\) to be classified.
The predicted class is the one with the highest similarity score. 

\subsection{Speaker-Specific Special Tokens}

We construct our experimental pipeline around speaker-dependent context modeling to facilitate a more natural handling of multi-speaker conversations. 
%Our approach can be divided into two stages: (1) adding speaker-specific special tokens to the tokenizer, and (2) constructing a context-aware dataset from raw conversational data.
For that, we create \emph{speaker tokens} for each character by converting the speaker name into an uppercase token and enclosing it in square brackets:
\begin{quote}
\texttt{[MONICA\_GELLER]}, \texttt{[ROSS\_GELLER]}, \dots, \texttt{[OTHER]}.
\end{quote}
All of these tokens are appended to the tokenizer as additional special tokens. This ensures that the transformer-based model can properly handle and learn representations for each specific speaker.

\subsection{Context-Aware Dataset Processing}
Our datasets contain a collection of multi-utterance scenes, each composed of multiple speaker turns. 
The processing begins by iterating over each scene and its corresponding utterances. 
For any given utterance, we extract the principal speaker and retrieve up to $\texttt{max\_previous\_context}$ previous utterances as conversational context.

Each previous utterance is prepended with its respective speaker token followed by the utterance text and a separator token.
We apply this procedure exclusively to the context utterances, as incorporating the speaker token of the target utterance would introduce an unfair advantage, thereby compromising the validity of the classification process.
The final input text for the current utterance is formed by:
\begin{enumerate}
    \item Adding the CLS token (\texttt{[CLS]}) at the beginning.
    \item Concatenating up to $\texttt{max\_previous\_context}$ previous utterances, each preceded by the corresponding speaker token and followed by the model's separator token (\texttt{[SEP]}).
    \item Appending the current utterance text.
    \item Adding a final separator token (\texttt{[SEP]}) at the end.
\end{enumerate}
Hence, every final utterance string that is passed to the model is of the form:
\begin{equation*}
\begin{array}{c}
\texttt{[CLS]} \;\; \texttt{[SPEAKER\_TOKEN]} \; \textit{utterance\_context} \;  \\ \texttt{[SEP]}
\;\dots\; \textit{current\_utterance} \;\; \texttt{[SEP]}.
\end{array}
\end{equation*}

%Through this method, we explicitly include speaker identity information in each utterance, enabling the model to capture speaker dynamics better and maintain contextual information across multi-turn dialogues.

%\subsection{Experimental Setup}
%\label{sec:exp_details}

%For our experiments, we employed the RoBERTa-base model. The training was conducted with a batch size of 32 and a learning rate of $1e^{-5}$. 
%The optimization process was carried out using the Adam optimizer across all training runs.

\begin{table*}[t!]
\centering
     \begin{tabular}{lccccccccccc} 
     \toprule
     & \multicolumn{7}{c}{\textbf{Class Individual F1}} & \\ \cmidrule{2-8}
     \textbf{Model} & \textbf{M} & \textbf{P} & \textbf{RA} & \textbf{RO} & \textbf{J} & \textbf{C} & \textbf{O} & \textbf{Accuracy}\\
     \midrule
     KNN & 13.30 & 12.13 & 17.34 & 19.23 & 14.68 & 14.61 & 19.23 & 16.18 \\ 
     RNN & 17.87 & 15.22 & 14.98 & 17.51 & 17.42 & 13.48 & 12.02 & 16.05 \\
     CNN & 20.55 & 17.52 & 24.20 & 24.70 & 28.15 & 14.05 & 31.81 & 25.01 \\
     Multi-Doc-CNN & 20.65 & 25.20 & 29.67 & 35.76 & 37.29 & 23.93 & 35.55 & 31.06 \\
     CNN-Concatenation  & 29.35 & 28.49 & 33.11 & 30.05 & 44.18 & 26.20 & 39.42 & 34.19 \\
     \midrule
     RoBERTa-Full (w/o token) & 33.62 & 34.34 & 41.75 & 47.80 & 45.13 & 32.73 & 44.13 & 40.74 \\
     %RoBERTa-FFP (w/o token) &  &  &  &  &  &  &  &  \\
     RoBERTa-Full (w/ token) & \textbf{67.63} & \textbf{75.01} & \textbf{73.37} & \textbf{69.16} & \textbf{71.66} & 67.59 & \textbf{68.27} & \textbf{70.56} \\
     RoBERTa-FFP (w/ token) & 65.29 & 73.05 & 68.82 & 65.72 & 70.67 & \textbf{67.99} & 67.58 & 68.74 \\
     \bottomrule
     %(w/ token) top-409
     \end{tabular}
\caption{Class individual F1 and Accuracy on \textit{Friends} Dataset. The first half contains baseline results from \cite{TEXT-SPEAKER-CNN}. The best RoBERTa-FFP has a fuzzy fingerprint $k=409$ (O - Other; RO - Ross; RA - Rachel; J - Joey; C - Chandler; P - Phoebe; M - Monica).}
\label{tab:friends-individual-results}
\end{table*}

%To investigate the effect of conversational context on speaker identification, we experimented with different context sizes, ranging from 0 (where only the speaker’s current turn is considered) to 5 (including the previous five turns). All experiments were performed on an NVIDIA RTX A6000 GPU with 48 GB of memory, enabling efficient processing of larger batch sizes and extended context windows.

%Model performance was evaluated using Accuracy, Macro F1, and Weighted F1, taking into account the unbalanced nature of the datasets.

\section{Results and Discussion}

\subsection{Comparison to Previous Approaches}

The approach from \cite{TEXT-SPEAKER-CNN} does not include any speaker-specific tokens to the context utterances, thus we also experiment with removing the speaker-specific tokens and only using the textual utterances (essentially, we remove \texttt{[SPEAKER\_TOKEN]} from the input).
Table~\ref{tab:friends-individual-results} shows that our model without the speaker-specific tokens already surpasses the best-performing baseline (CNN-Concatenation) by over 6 points in terms of accuracy (40.74\% vs.\ 34.19\%). %Furthermore, our best model achieves significant gains over all baselines when we incorporate the speaker tokens.

However, when we include the speaker tokens, the performance improves substantially across all classes, reaching an overall accuracy of 70.56\%. 
This outperforms all previous baselines by a large margin, confirming that speaker-specific embeddings are critical for capturing the nuances of each character in the dialogue. 
Notably, the gains are consistent across every class, indicating that the speaker embeddings allow the model to learn character-specific language patterns more effectively.
These results highlight the importance of modeling speaker identity alongside textual utterances. While conventional text-only methods do provide a reasonable signal for classification, explicitly encoding the speaker tokens helps the model better disambiguate similar dialogue styles and linguistic cues unique to each character. 

%In particular, multi-party conversations often exhibit frequent speaker changes, back-and-forth exchanges, and overlapping references; these patterns are closely tied to the identity of the speaker. By incorporating speaker tokens, the model becomes aware of typical dialogue dynamics: for instance, that a new turn usually shifts to a different speaker or that two speakers might be engaged in a rapid back-and-forth. This contextual awareness allows the model to leverage turn-taking structure and speaker-specific language usage more effectively. In line with our initial hypothesis, these results demonstrate that in multi-party settings, the conversational context cannot be fully captured through text alone: explicitly modeling each speaker is crucial to achieving more accurate and robust performance.

\subsection{Fuzzy Fingerprint Size Variation}

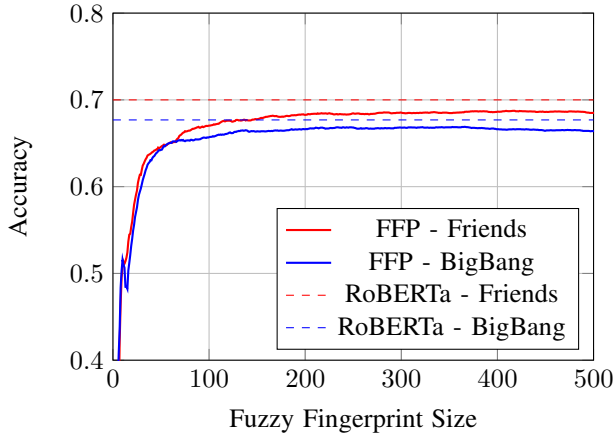
\begin{figure}[ht]
\centering
\begin{tikzpicture}
\begin{axis}[
    width=0.9\linewidth,
    height=0.7\linewidth,
    xlabel={Fuzzy Fingerprint Size},
    ylabel={Accuracy},
    xmin=0, xmax=500,
    ymin=0.4, ymax=0.8,
    legend pos=south east,
    grid=major,
    %grid style={dotted},
    % title={Accuracy, Macro F1, and Weighted F1 vs. Fingerprint Size}, % Optional
]
\addplot[
    color=red,
    %mark=square*,
    thick
] table[x=fingerprint_size, y=accuracy, col sep=comma] {friends_fingerprints.csv};
\addlegendentry{FFP - Friends}
\addplot[
    color=blue,
    thick
    %mark=square*,
    %thick
] table[x=fingerprint_size, y=accuracy, col sep=comma] {bigbang_fingerprints.csv};
\addlegendentry{FFP - BigBang}
\addplot[
    color=red,
    domain=0:500,
    %thick,
    dashed
]{0.70};
\addlegendentry{RoBERTa - Friends}
\addplot[
    color=blue,
    domain=0:500,
    %thick,
    dashed
]{0.6769};
\addlegendentry{RoBERTa - BigBang}

\end{axis}
\end{tikzpicture}
\caption{Accuracy variation with different fuzzy fingerprint sizes on the \textit{Friends} and \textit{Big Bang Theory} datasets. The Fuzzy Fingerprint (FFP) model (solid lines) retains only the top-$k$ hidden units from the last hidden layer, while RoBERTa (dashed lines) utilizes all 768 hidden units.}
\label{fig:fingerprint_variation}
\end{figure}

Figure~\ref{fig:fingerprint_variation} shows how the performance of our Fuzzy Fingerprint framework on the \textit{Friends} and \textit{Big Bang Theory} datasets varies with the size of the fuzzy fingerprint (i.e., the top-$k$ hidden units retained from the last hidden layer). 
As we increase $k$, the accuracy rapidly improves and approaches the score achieved by the fully fine-tuned RoBERTa model (which leverages all 768 units in its last hidden layer). Notably, beyond $k \approx 150$, the FFP curve saturates slightly below RoBERTa, indicating that retaining only a fraction of the hidden units is sufficient to achieve nearly the same level of accuracy as using all 768 dimensions.

%These results suggest further exploration of pruning techniques in transformer-based models. Since it is possible to approach RoBERTa’s performance with a reduced number of hidden units, a structured pruning method could be employed to identify and remove redundant dimensions, yielding a more lightweight and efficient model.

Another interesting benefit of the fuzzy fingerprint representation is its direct interpretability. By maintaining class-specific fuzzy fingerprints, we can measure a sample’s similarity to each class fuzzy fingerprint using Equation \ref{eq:similarity}. 
This property of fuzzy fingerprints enables analyzing how a given sample’s fuzzy fingerprint aligns (or fails to align) with a specific speaker's fuzzy fingerprint, revealing when the model is most uncertain. 
In such cases, the utterance might be considered generic, meaning it lacks strong stylistic cues and could plausibly belong to multiple speakers.

\subsection{Influence of the Context}
\label{subsec:influence_of_context}

% 366 BIG BANG

\begin{table*}[t]
\centering
     \begin{tabular}{lccccccc} 
     \toprule
     & \multicolumn{3}{c}{\textbf{Friends Corpus}} & \multicolumn{3}{c}{\textbf{Big Bang Theory Corpus}} \\
     \cmidrule(lr){2-4} \cmidrule(lr){5-7}
     \textbf{\# Context} & \textbf{M-F1} & \textbf{W-F1} & \textbf{Accuracy (\%)} & \textbf{M-F1} & \textbf{W-F1} & \textbf{Accuracy (\%)} \\
     \midrule
     0 & 26.65 & 27.25 & 27.11 & 23.63 & 29.40 & 32.28 \\
     1 & 36.87 & 37.36 & 37.35 & 37.91 & 42.77 & 44.60 \\
     2 & 67.80 & 68.13 & 68.06 & 63.54 & 65.11 & 65.16 \\
     3 & 69.54 & 69.84 & 69.81 & 63.94 & 66.05 & 66.35 \\
     4 & 70.32 & 70.62 & \textbf{70.63} & 64.85 & 66.95 & 67.09 \\
     5 & \textbf{70.39} & \textbf{70.64} & 70.56 & \textbf{65.45} & \textbf{67.41} & \textbf{67.69} \\
     6 & 69.74 & 69.97 & 69.98 & 64.37 & 66.28 & 66.52 \\
     \bottomrule
     \end{tabular}
\caption{Performance for different context sizes across \textit{Friends} and \textit{Big Bang Theory} corpora (M-F1 - Macro F1, W-F1 - Weighted F1).}
\label{tab:context-comparison}
\end{table*}

As we observe in Table~\ref{tab:context-comparison}, performance steadily increases as the number of prior utterances (i.e., the context) grows from 0 to 5 in both the \textit{Friends} and \textit{Big Bang Theory} corpora. This trend is intuitive, given that adding more conversational turns supplies richer cues regarding interlocutors, ongoing topics, and preceding statements. In a dialogue setting, however, many utterances are responses to immediate questions, references to a previous statement, or brief interjections. When the model lacks contextual information, it often fails to grasp the question-and-answer flow or to detect back-and-forth interactions that are essential for identifying the speaker. By including more turns, the model situates each utterance within a more detailed exchange, thereby achieving markedly higher scores: for instance, \textit{Friends}’ accuracy rises from 27.11\% (no context) to over 70\% when five preceding turns are available.
Nonetheless, adding too many utterances (e.g., \(\#\)~Context\,=\,6) can slightly diminish performance in both corpora, likely due to older context becoming less relevant or introducing extraneous details. %This subtle drop underscores the balanced nature of dialogue data: while some recent turns are highly informative, excessive context may dilute the model’s focus, particularly in shorter scenes where older utterances have less bearing on the current exchange. Thus, carefully selecting how many previous turns to incorporate helps maintain an optimal balance between relevance and noise. 
%Overall, these findings highlight the crucial role of context in speaker identification, particularly in conversation-heavy domains, and reveal how effectively leveraging a modest window of prior utterances can substantially boost classification accuracy.

\subsection{Example Analysis}
\label{subsec:examples}

\begin{figure}[ht]
\centering
\begin{tikzpicture}
    \begin{axis}[
        width=8cm,
        height=6cm,
        ymin=0,
        ymax=0.13,
        xmin=0,
        xmax=25,
        ybar interval,
        bar width=5,
        scaled y ticks=false,  % Disables scientific notation
        yticklabel style={/pgf/number format/fixed},  % Forces decimal notation
        xlabel={Length of Utterances (Number of Words)},
        ylabel={Relative Frequency},
        legend pos=north east,
        grid=major,
        xtick=\empty
    ]
    \addplot+[
        ybar,
        fill=blue,
        opacity=0.7
    ] table [x index=0, y index=2, col sep=comma] {utterance_length_histogram.csv};
    \addlegendentry{Correct Samples}

    \addplot+[
        ybar,
        fill=red,
        opacity=0.7
    ] table [x index=0, y index=3, col sep=comma] {utterance_length_histogram.csv};
    \addlegendentry{Incorrect Samples}

    \end{axis}
\end{tikzpicture}
\caption{Histogram of utterance lengths comparing correct and incorrect predictions. Darker red areas indicate regions where correct and incorrect samples overlap.}
\label{fig:histogram_correct_incorrect}
\end{figure}
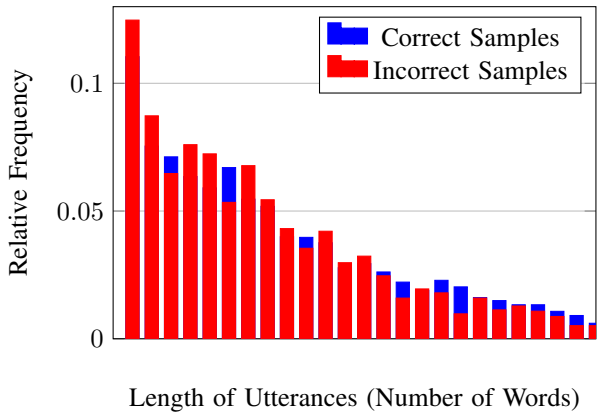

%To further illustrate the model's behavior, we present a qualitative analysis of two representative examples from the \textit{Friends} corpus. In each instance, we report the model's predicted label, the ground-truth speaker, and the computed fuzzy similarities (Equation~\ref{eq:similarity}) to each candidate class.

\begin{itemize}[leftmargin=*]
    \item \textit{Example 1:}
    \\
    \verb|[CLS] Great! Just give me a sec to | \\
    \verb|change film. [SEP]|
    \smallskip
    
    \noindent
    \textbf{Similarity scores:} 
    \textit{Monica Geller} (0.3537), 
    \textit{Ross Geller} (0.3561), 
    \textit{Joey Tribbiani} (0.3649), 
    \textit{Phoebe Buffay} (0.3354), 
    \textit{Rachel Green} (0.3426), 
    \textit{Chandler Bing} (0.3454), 
    \textit{Other} (0.3601).
    \\
    \textbf{Correct Label:} \textit{Other} \quad
    \textbf{Predicted Label:} \textit{Joey Tribbiani}
    \smallskip

    \noindent
    Here, the semantic content of the utterance (\textit{“Just give me a sec to change film.”}) does not strongly indicate any particular main character, which could suggest that it is generic to any speaker. This confusion reflects the difficulty in distinguishing non-core speakers from main characters when the utterance lacks strong persona cues or is too brief to leverage contextual patterns effectively.

    %\bigskip
    %\item \textit{Example 2:}
    %\\
    %\verb|[CLS] [JOEY_TRIBBIANI] Hey. [SEP] Hey.|
    %\verb|[SEP]|
    %\smallskip

    %\noindent
    %\textbf{Similarity scores:} 
    %\textit{Monica Geller} (0.3000),
    %\textit{Ross Geller} (0.3187), 
    %\textit{Joey Tribbiani} (0.2943), 
    %\textit{Phoebe Buffay} (0.2716), 
    %\textit{Rachel Green} (0.2754), 
    %\textit{Chandler Bing} (0.3250), 
    %\textit{Other} (0.2906).
    %\\
    %\textbf{Correct Label:} \textit{Ross Geller} \quad
    %\textbf{Predicted Label:} \textit{Chandler Bing}
    %\smallskip
    
    %\noindent
    %Here, the model mistakenly assigns the label \textit{Chandler Bing}, likely due to the colloquial greeting ``Hey.'' that often appears across multiple characters. The similarity values are all relatively close, underscoring how short, generic utterances can produce confusion when speaker-specific linguistic traits are minimal. 

    \bigskip
    \item \textit{Example 2:}
    \\
    \verb|[CLS] [JOEY_TRIBBIANI] I would, but this |    \\
    \verb|is a nice place and my T-shirt... [SEP] |    \\
    \verb|[RACHEL_GREEN] Oh my God! Really?! Can |    \\
    \verb|I see it? [SEP] Yeah. Sure. [SEP]|
    \smallskip

    \noindent
    \textbf{Similarity scores:} 
    \textit{Monica Geller} (0.3457), 
    \textit{Ross Geller} (0.3398), 
    \textit{Joey Tribbiani} (0.4546), 
    \textit{Phoebe Buffay} (0.3100), 
    \textit{Rachel Green} (0.3137), 
    \textit{Chandler Bing} (0.3449), 
    \textit{Other} (0.3244)
    \\
    \textbf{Correct Label:} \textit{Joey Tribbiani} \quad
    \textbf{Predicted Label:} \textit{Joey Tribbiani}
    \smallskip

    \noindent
    In contrast to the first example, this utterance carries strong persona cues (omitted in the example, but Joey mentions a humorous T-shirt and casual banter, which align well with Joey's persona). The model confidently picks \textit{Joey Tribbiani}, with a higher similarity score (0.4546) than any other character. This success illustrates the importance of context and personal references in guiding the prediction. 
    The presence of the speaker token \texttt{[JOEY\_TRIBBIANI]} in earlier turns, combined with Joey’s characteristic comedic style, helps the model correctly maintain speaker consistency.
    \smallskip

\end{itemize}

We also explore the correlation between the size of the utterances and the number of correct/incorrect examples.
As indicated by Figure~\ref{fig:histogram_correct_incorrect}, shorter utterances are more likely to be misclassified compared to their longer counterparts. The histogram shows a clear concentration of incorrect predictions (red bars) in the shortest utterance bins, while longer utterances have a higher proportion of correct classifications (blue bars). 
%This pattern suggests that brief statements often lack strong speaker-specific markers, making them more susceptible to ambiguity. 
Many short utterances in dialogue are generic responses such as acknowledgments, interjections, or simple affirmations, which multiple characters could plausibly say. 
Conversely, longer utterances tend to contain more distinctive phrasing, speaker idiosyncrasies, or contextual clues that anchor them to a specific character. This underscores the importance of incorporating additional conversational context or auxiliary speaker information to mitigate ambiguity in short utterances.

\def\myConfMat{{
{541,  28,   54,  29,  64,  38,  59},
 { 26, 705,  64,  15,  79, 37,  63},
 { 55,  39, 628,  22,  76,  32,  57},
 { 83,  36,  36, 440,  80,  36,  63},
 { 55,  54,  36,  23, 836,  13,  71},
 { 40,  14,  55,  17,  53, 446,  52},
 { 44,  65,  43,  19,  90,  33, 687},
}}

\def\classNames{{"M","RO","J","P","RA", "C", "O"}} %class names. Adapt at will

\def\numClasses{7} %number of classes. Could be automatic, but you can change it for tests.

\def\myScale{0.8} % 1.5 is a good scale. Values under 1 may need smaller fonts!
\begin{figure}[ht]
    \centering
    \begin{tikzpicture}[
        scale = \myScale,
        %font={\scriptsize}, %for smaller scales, even \tiny may be useful
        ]
    
    \tikzset{vertical label/.style={rotate=90,anchor=east}}   % usable styles for below
    \tikzset{diagonal label/.style={rotate=45,anchor=north east}}
    
    \foreach \y in {1,...,\numClasses} %loop vertical starting on top
    {
        % Add class name on the left
        \node [anchor=east] at (0.4,-\y) {\pgfmathparse{\classNames[\y-1]}\pgfmathresult}; 
        
        \foreach \x in {1,...,\numClasses}  %loop horizontal starting on left
        {
    %---- Start of automatic calculation of totSamples for the column ------------   
        \def\totSamples{0}
        \foreach \ll in {1,...,\numClasses}
        {
            \pgfmathparse{\myConfMat[\ll-1][\x-1]}   %fetch next element
            \xdef\totSamples{\totSamples+\pgfmathresult} %accumulate it with previous sum
            %must use \xdef fro global effect otherwise lost in foreach loop!
        }
        \pgfmathparse{\totSamples} \xdef\totSamples{\pgfmathresult}  % put the final sum in variable
    %---- End of automatic calculation of totSamples ----------------
        
        \begin{scope}[shift={(\x,-\y)}]
            \def\mVal{\myConfMat[\y-1][\x-1]} % The value at index y,x (-1 because of zero indexing)
            \pgfmathtruncatemacro{\r}{\mVal}   %
            \pgfmathtruncatemacro{\p}{round(\r/\totSamples*100)}
            \coordinate (C) at (0,0);
            \ifthenelse{\p<50}{\def\txtcol{black}}{\def\txtcol{white}} %decide text color for contrast
            \node[
                draw,                 %draw lines
                text=\txtcol,         %text color (automatic for better contrast)
                align=center,         %align text inside cells (also for wrapping)
                fill=red!\p,        %intensity of fill (can change base color)
                minimum size=\myScale*10mm,    %cell size to fit the scale and integer dimensions (in cm)
                inner sep=0,          %remove all inner gaps to save space in small scales
                ] (C) {\p\%};    %\\\p\%  %text to put in cell (adapt at will)
            %Now if last vertical class add its label at the bottom
            \ifthenelse{\y=\numClasses}{
            \node [] at ($(C)-(0,0.75)$) % can use vertical or diagonal label as option
            {\pgfmathparse{\classNames[\x-1]}\pgfmathresult};}{}
        \end{scope}
        }
    }
    %Now add x and y labels on suitable coordinates
    \coordinate (yaxis) at (-0.8,0.5-\numClasses/2);  %must adapt if class labels are wider!
    \coordinate (xaxis) at (0.5+\numClasses/2, -\numClasses-1.3); %id. for non horizontal labels!
    \node [vertical label] at (yaxis) {Correct Class};
    \node []               at (xaxis) {Predicted Class};
    \end{tikzpicture}
\caption{Confusion matrix for the \textit{Friends} dataset using a fuzzy fingerprint $k=409$.}
\label{fig:confusion_matrix}
\end{figure}

The confusion matrix in Figure~\ref{fig:confusion_matrix} reveals that every character is misclassified as nearly every other character to some extent, highlighting the challenge of distinguishing speakers in multi-party conversations. While the diagonal values indicate strong overall performance, the off-diagonal misclassifications suggest that certain utterances lack clear persona cues, making them prone to ambiguity. This pattern aligns with our example analysis, where generic utterances, those that could plausibly be spoken by multiple characters, were frequently misclassified. 
%Notably, \textit{Other} (O) is often assigned to main characters, reinforcing the idea that generic speech tends to be absorbed into dominant speaker patterns.

\subsection{Capturing Generic Utterances}

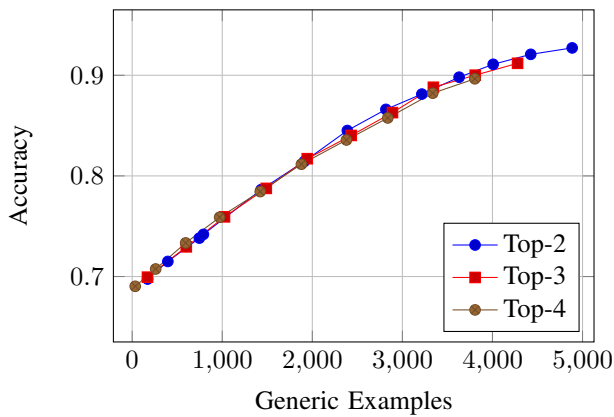
\begin{figure}[ht]
    \centering
    \begin{tikzpicture}
    \begin{axis}[
        xlabel={Generic Examples},
        ylabel={Accuracy},
        legend pos=south east,
        width=0.9\columnwidth,
        height=6cm,
        grid=major,
        ymin=0.65, ymax=0.95,
        enlargelimits=0.05
    ]
    \addplot table[x=Top2_Removed, y=Top2_Accuracy, col sep=semicolon, /pgf/number format/read comma as period] {generic_test.csv};
    \addlegendentry{Top-2}
    
    \addplot table[x=Top3_Removed, y=Top3_Accuracy, col sep=semicolon, /pgf/number format/read comma as period] {generic_test.csv};
    \addlegendentry{Top-3}
    
    \addplot table[x=Top4_Removed, y=Top4_Accuracy, col sep=semicolon, /pgf/number format/read comma as period] {generic_test.csv};
    \addlegendentry{Top-4}
    \end{axis}
    \end{tikzpicture}
    \caption{Accuracy variation as generic utterances are removed, where Top-2, Top-3, and Top-4 denote cases where an utterance has similar scores for two, three, or four speaker fuzzy fingerprints, respectively.}
    \label{fig:generic_examples}
\end{figure}

%As we have seen, in the Fuzzy Fingerprint framework, each sample’s fingerprint is compared to all class fingerprints (Equation~\ref{eq:similarity}), where the class with the highest similarity is selected as the predicted label. However, as illustrated in our example analysis, certain utterances, particularly short or generic ones, exhibit very close similarity scores across multiple classes, making classification ambiguous. In such cases, assigning a single label may introduce artificial certainty where none exists.

To better capture the uncertainty, we explore applying different thresholds based on the similarities obtained from the top-scoring speakers. 
Specifically, we compare the similarity of an utterance to its top-2, top-3, and top-4 most similar class fuzzy fingerprints. The intuition behind this approach is straightforward: an utterance may be considered generic if it could plausibly belong to multiple speakers rather than being uniquely attributed to one. By varying this threshold, we assess how different levels of tolerance for ambiguity affect classification performance.

Figure~\ref{fig:generic_examples} presents the results of this experiment. The curves correspond to considering the top speakers, where we progressively filter out utterances classified as generic. 
As more ambiguous examples are excluded, accuracy steadily improves, confirming that many misclassifications arise from borderline cases where multiple speakers exhibit similar linguistic patterns. These findings suggest that future work could explore adaptive thresholds or context-aware techniques to dynamically assess speaker ambiguity and improve robustness in multi-party dialogue settings.

\section{Conclusions and Future Work}

In this paper, we investigated text-based speaker identification by exploring fuzzy fingerprints from large pre-trained models, speaker-specific tokens, and context-aware modeling. Our findings show that incorporating conversational context significantly improves classification accuracy, with optimal context sizes around three to five utterances. Additionally, we demonstrated that fuzzy fingerprints can approximate full fine-tuning performance with fewer hidden units, providing a more interpretable alternative.

Despite these advancements, short and generic utterances remain a key challenge, often leading to ambiguous classifications. Future research should focus on refining context modeling, improving speaker disambiguation, and integrating additional textual cues to enhance robustness. %By making our datasets and code publicly available, we aim to facilitate further advancements in text-based speaker identification.

\bibliographystyle{IEEEtran}
\bibliography{references}

\end{document}